\documentclass{article}

\usepackage[dblblindworkshop, final]{neurips_2025}

\usepackage[utf8]{inputenc} 
\usepackage[T1]{fontenc}    
\usepackage{hyperref}       
\usepackage{url}            
\usepackage{booktabs}       
\usepackage{amsfonts}       
\usepackage{nicefrac}       
\usepackage{microtype}      
\usepackage{xcolor}         
\usepackage{enumitem}
\usepackage{graphicx}
\usepackage{amsmath}
\usepackage{verbatim}
\usepackage[most]{tcolorbox}

\title{OrchDAG: Complex Tool Orchestration in Multi-Turn Interactions with Plan DAGs}

\author{%
  Yifu Lu\thanks{Work is done during internship in Amazon. Equal contribution.} \\
  Princeton University\\
  \texttt{yiful@princeton.edu} \\
  \And
  Shengjie Liu\footnotemark[1] \\
  Amazon \\
  \texttt{zycjlsj@amazon.com} \\
  \AND
  Li Dong \\
  Amazon \\
  \texttt{ldonga@amazon.com} \\
}

\begin{document}

\maketitle

\begin{abstract}
 Agentic tool use has gained traction with the rise of agentic tool calling, yet most existing work overlooks the complexity of multi-turn tool interactions. We introduce OrchDAG, a synthetic data generation pipeline that models tool execution as directed acyclic graphs (DAGs) with controllable complexity. Using this dataset, we benchmark model performance and propose a graph-based reward to enhance RLVR training. Experiments show that the dataset presents a challenging but solvable benchmark, and the proposed reward is effective when combined with GRPO-style algorithms, highlighting the importance of leveraging topological structure and data complexity in multi-turn tool use.
\end{abstract}

\section{INTRODUCTION}
\label{intro}


Large Language Models (LLMs) (\cite{brown2020languagemodelsfewshotlearners, JMLR:v24:22-1144,deepseekai2025deepseekr1incentivizingreasoningcapability,touvron2023llama2openfoundation,zeng2023glm130bopenbilingualpretrained}) have been at the forefront of advancing artificial intelligence, marking significant breakthroughs in diverse fields. The planning capabilities of LLMs, particularly their ability to use tools (\cite{yao2023treethoughtsdeliberateproblem, yao2023reactsynergizingreasoningacting}), enable them not only to execute commands and perform web searches but also to enhance their advanced mathematical reasoning abilities. LLM Compiler \cite{kim2024llmcompilerparallelfunction} and its subsequent work (\cite{erdogan2024tinyagentfunctioncallingedge, erdogan2025planandactimprovingplanningagents}) propose constructing tooling usage as a directed acyclic graph (DAG) to enable the parallel execution of independent tools, thereby improving tool-calling efficiency. CodeAct \cite{wang2024executablecodeactionselicit} and CodePlan \cite{wen2024unlockingreasoningpotentiallarge} propose leveraging the generation of pseudo-Python code to outline high-level reasoning processes for complex multi-step reasoning tasks, where each tool usage is represented as a function call within the code. ReWOO \cite{xu2023rewoodecouplingreasoningobservations} proposes a modular framework that decouples the reasoning process from the external observations of each tool usage, thereby reducing token consumption and improving efficiency. \cite{liu2025toolplannertaskplanningclusters} clusters the provided tools into groups of toolkits, plans at the toolkit level, and replans by selecting tools within the same toolkit if error comes out. \cite{ma2024nonmyopicgenerationlanguagemodels} proposes a method called Predictive-Decoding, which leverages Model Predictive Control from the optimal control field to mitigate early errors in planning and promote non-myopic planning, thereby enhancing overall accuracy. ReasonFlux \cite{yang2025reasonfluxhierarchicalllmreasoning} proposes a framework in which the LLM reasons over template fields, executes tools based on the templates, and employs reinforcement learning to improve planning accuracy using an action completion reward.

In the agentic setting, LLMs are evolving beyond purely textual reasoning toward dynamic agents capable of planning, tool use, and multi-step (also multi-turn) execution. The introduction of Group Relative Policy Optimization (GRPO)  \cite{shao2024deepseekmathpushinglimitsmathematical} further inspired the development of Reinforcement Learning with Verifiable Reward (RLVR) for agentic tool use, driven by its efficiency. The xLAM \cite{zhang2024xlamfamilylargeaction} suite introduces purpose-built “large action models” optimized for function calling, offering strong baselines and open resources for multi-turn tool execution. Llama-Nemotron \cite{bercovich2025llamanemotronefficientreasoningmodels} extends this trajectory with efficient reasoning modes and scalable inference, enabling models to dynamically switch behaviors across long conversations. ToolRL \cite{qian2025toolrlrewardtoollearning} systematically studies reinforcement learning reward designs—covering granularity, temporal structure, and signal types—to improve generalization in multi-turn tool-integrated reasoning, while OTC \cite{wang2025actingreasoningmoreteaching} complements this by explicitly balancing accuracy and tool-call efficiency to maintain productivity over prolonged interactions. Kimi K2 \cite{kimiteam2025kimik2openagentic} shows that stabilizing long-context training and using multi-stage RL leads to robust performance across multi-round software engineering, math, and agentic tasks. These works highlight that advancing agentic LLMs in multi-turn settings requires not only larger or more efficient models, but also principled reward design, cost-aware tool-use strategies, and scalable system pipelines.

Recent advances in evaluating agentic models have led to new benchmarks and environments for assessing performance in realistic, interactive scenarios. ACEBench \cite{chen2025acebenchwinsmatchpoint} overcomes limits of prior evaluations by introducing a structured benchmark with Normal, Special, and Agent categories to test atomic-level tool use across simple, complex, multi-agent, and ambiguous instruction settings. Complementing this, BFCL (v3) \cite{patil2025bfcl} standardizes function-calling benchmarks across real-world contexts, supporting serial and parallel invocations in multiple languages through an AST-based evaluation. $\tau$-Bench, $\tau^2$-Bench, and UserBench \cite{yao2024taubenchbenchmarktoolagentuserinteraction, barres2025tau2benchevaluatingconversationalagents, qian2025userbenchinteractivegymenvironment} together extend evaluation from structured tool–user interactions to controlled bidirectional agent cooperation and finally to fully user-centric, dynamic environments, progressively enriching the realism and robustness of agent assessment.

Prior work on tool use has mainly studied general real-world APIs \cite{qin2023toolllmfacilitatinglargelanguage}, such as \textit{send email} or \textit{make calendar}, along with related functions in \textit{web search} systems like Manus. Meanwhile, current multi-turn settings mainly focus on computer-use tasks \cite{patil2025bfcl}, such as manipulating files in the operating system. In industrial settings, however, an agent may need to work with hundreds of domain tools, including APIs and pipeline endpoints, and may also interact with other domain agents to produce a complete answer to a user query. The complexity usually arises from three aspects: (1) the dependencies among tools can be intricate, (2) the output of a tool is often represented as a JSON file with many fields, and (3) a key output field from one tool may serve as an input to another, but with different field names. Moreover, in multi-turn settings, the environment may execute the required tools but encounter time-outs or runtime errors in their responses. Therefore, it is important to construct a dataset that not only evaluates current models but also pushes their capabilities in complex multi-turn tool interaction, which is essential for building robust and reliable agents.

In summary, this work makes several pivotal contributions:
\begin{itemize}[left=0pt, itemsep=0pt]
    \item We design a synthetic multi-turn data generation pipeline OrchDAG for agentic tool use, where each round of tool execution for a user query is represented as a DAG. The complexity of the generated data is controlled by a pipeline hyperparameter. 
    \item Using the constructed dataset, we first evaluate the current model's performance and then introduce a graph-based reward derived from the DAG for RLVR training. 
    \item Extensive experiments show the effectiveness of our approach, emphasizing the value of incorporating the topological structure of tool execution graphs and the importance of controlling data complexity in multi-turn tool use. 
\end{itemize}

\section{PRELIMINARY}

\subsection{LLM Reasoning with Tools for Multi-turn Settings}
For the first turn, given a user query $x$ and a pretrained LLM $\rho_{\theta}(\cdot)$, the LLM generates an tool execution plan represented as a graph with $p = \{\mathcal{P}_1, \dots, \mathcal{P}_n\} \sim \rho_{\theta}(p \mid \mathcal{T}, \mathcal{D}, x)$, where $p$ is the plan list after topological sorting, $\mathcal{T}$ is the set of available tools, and $\mathcal{D}$ is the collection of descriptions for all available tools. At each step $t$, the LLM generates an intermediate reasoning output $r_t \sim \rho_{\theta}(r_t \mid \mathcal{T}, \mathcal{D}, x, p, \mathcal{O}_1, \dots, \mathcal{O}_{t-1})$ and executes the plan step $\mathcal{P}_t$ to obtain the observation $\mathcal{O}_t$. The final response is then generated as $\mathcal{R} \sim \rho_{\theta}(\mathcal{R} \mid \mathcal{T}, \mathcal{D}, x, p, \mathcal{O}_1, \dots, \mathcal{O}_n)$. 

In later turns, a user may issue an irrelevant query requiring a completely new tool execution graph, or a dependent query that builds on partial outputs from earlier tool executions or responses. Additionally, some tools may return errors (e.g., timeouts), requiring unfinished execution paths from prior queries to be rescheduled.

\subsection{Tool Execution as DAG}
Given a plan $p$ generated by the LLM, we represent it as a directed graph $\mathcal{G} = (\mathcal{V}, \mathcal{E})$, where $\mathcal{V} = (v_1, \dots, v_n)$ is the set of nodes and $\mathcal{E} = (e_1, \dots, e_m)$ is the set of edges. The node $v_1$ corresponds to the user query, and $v_n$ represents the final node that aggregates observations and returns the response. The intermediate nodes $v_2, \dots, v_{n-1}$ correspond to tool calls, each associated with an attribute that stores its tool payload in JSON format. An edge $e_i \in \mathcal{E}$ denotes a dependency between two tools, where an output key from the source tool serves as an input key to the target tool.

We represent the tool graph as an ordered list of tasks in a JSON-like text style. Each task contains four fields \label{task_list}: \texttt{task\_id}, \texttt{toolname}, \texttt{payload}, and \texttt{dependencies}. A task can be associated with multiple dependencies. A task can be expressed as \{\texttt{task\_id}: \texttt{task\_4}, \texttt{toolname}: \texttt{name}, \texttt{payload}: \{\texttt{param1}: \texttt{val1}, \texttt{param2}: \texttt{\$2.outputkey1}, \texttt{param3}: \texttt{\$3.outputkey4}\}, \texttt{dependencies}: [\texttt{task\_2}, \texttt{task\_3}]\}.

\section{METHODOLOGY}

In real-world domains, API specifications and orchestrations are often considerably more complex, making it challenging for LLMs pretrained on general public data to generate accurate and reliable plans for diverse user queries. Drawing inspiration from LLMCompiler \cite{kim2024llmcompilerparallelfunction, erdogan2025planandactimprovingplanningagents}, for queries involving complex tool interactions, it is advantageous to first construct a tool execution DAG. This DAG serves as a blueprint for executing tools sequentially or in parallel, with subsequent replanning guided by both the execution results of the current DAG and any new user queries in later turns. 

\subsection{OrchDAG – Synthetic Data Generation Pipeline for Multi-Turn Tool Use}

As discussed in Section~\ref{intro}, the design of the data generation pipeline follows several key principles to better reflect real-world tool-use scenarios: (1) the complexity of the tool execution DAG for each synthetic user query should be controllable through pipeline hyperparameters, (2) the system prompt provided to the LLM must include irrelevant tools (both schema and description) so that the model learns to identify and select only the relevant ones, (3) the output payload of each tool should contain multiple fields, typically four or five, and (4) at least one key output field from a tool should serve as an input to another tool, but with a different field name, to capture schema misalignment commonly observed in practice.

Moreover, in multi-turn settings, the data should capture scenarios where certain nodes in the tool execution DAG fail. When the user issues a follow-up query, the corresponding DAG in the dataset should exclude nodes that have already been executed in the previous turn, while reusing their available results whenever applicable. In light of these requirements, we develop a graph-based data generation pipeline implemented with LangGraph \footnote{https://langchain-ai.github.io/langgraph/}, accompanied by a set of validation functions to ensure the quality of the generated data points.

\begin{figure}[h]
  \centering
  \includegraphics[width=\linewidth]{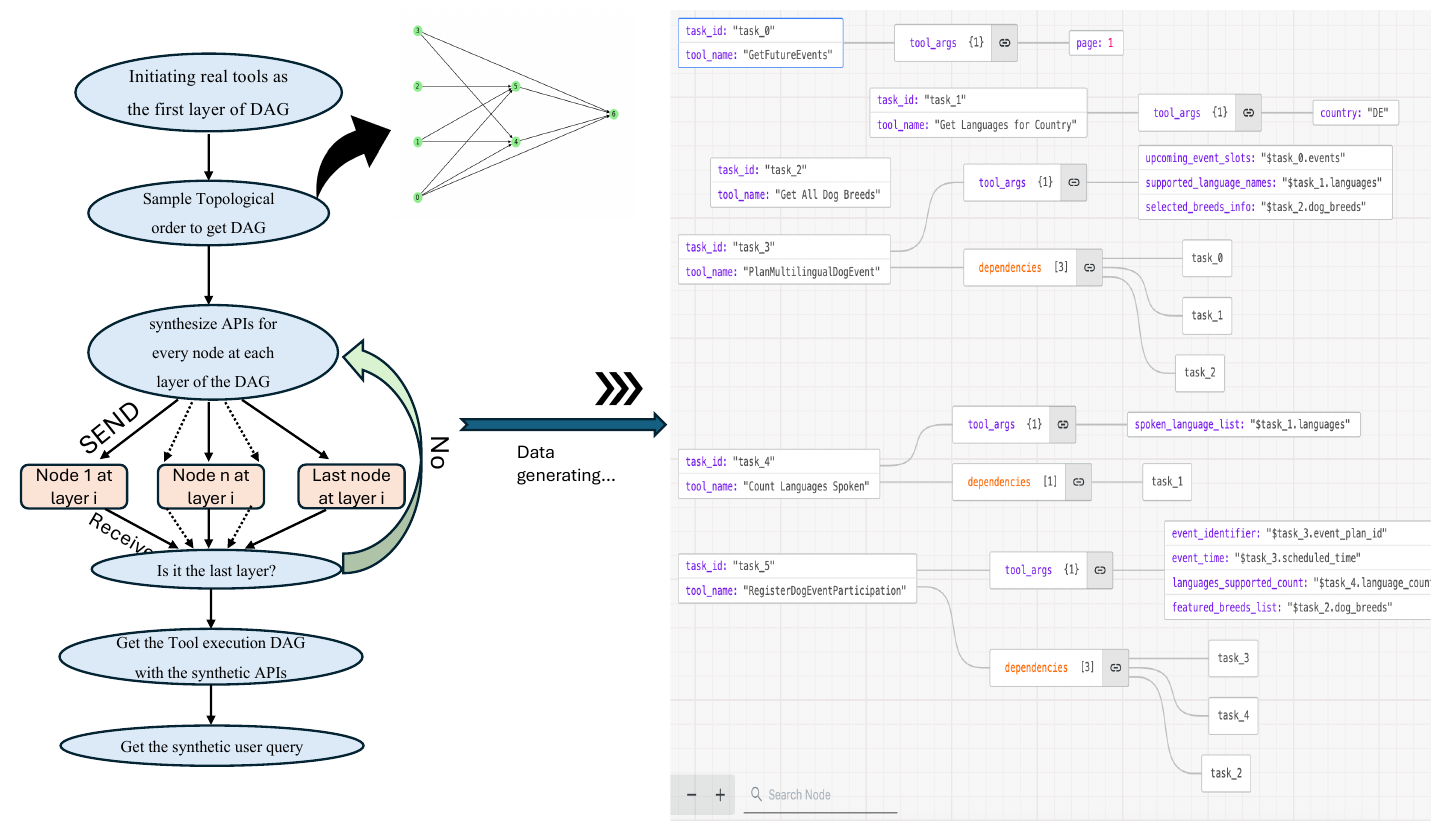} 
  \caption{Data Generation Pipeline for a single turn}
  \label{fig:image1}
\end{figure}

As shown in Figure \ref{fig:image1}, the data generation pipeline begins by collecting real high-quality tools from existing benchmarks, Specifically, we leverage \textsc{APIGen} \cite{liu2024apigenautomatedpipelinegenerating} and \textsc{ToolACE} \cite{liu2025toolacewinningpointsllm}. To ensure sufficient complexity, we retain only examples where the final answer involves more than two distinct functions represented in JSON format. Notably, we extract tools directly from the answers rather than from the system-prompt tool lists. This design choice eliminates the need for additional categorization or clustering steps, which could introduce unnecessary uncertainty, while naturally yielding a smaller and more coherent set of related tools. After filtering, \textsc{APIGen} contributes 2,542 data points and \textsc{ToolACE} contributes 1,005. \label{datapoint}

For each data point, suppose it contains four real tools; these are placed as the first layer of the tool execution DAG. Based on the hyperparameters of the DAG (height and width), we then randomly sample a topological order to obtain a DAG template (see Figure~\ref{fig:image1}). According to this template, we synthesize the tools for each node layer by layer under the following conditions: (1) each input key must depend on an output key from one of its parent nodes in the DAG, and (2) the field names should not remain identical across instances but instead vary randomly. After populating the DAG template, we obtain the tool execution DAG, which is then used to prompt the LLM to generate the corresponding user query, conditioned on the DAG and a few-shot set of examples. Finally, we augment the system prompt with irrelevant tools to encourage the model to discriminate among available options.

\begin{figure}[ht]
  \centering
  \includegraphics[width=\linewidth]{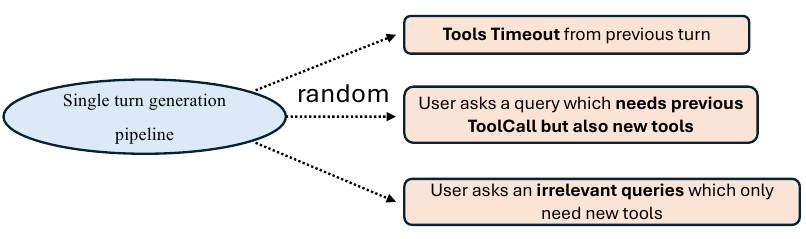} 
  \caption{Extension of single-turn data generation pipeline to multi-turn settings}
  \label{fig:image2}
\end{figure}

We further extend the generation pipeline to the multi-turn setting by attaching three additional nodes to the final node (see Figure \ref{fig:image2}), with only one being activated during generation. These nodes correspond to three possible multi-turn scenarios \label{multiturn}: (1) the user issues a completely irrelevant query, (2) the user poses a query that requires a new set of tools while also depending on the previous final response or intermediate tool outputs, and (3) a tool execution error occurs, such as a timeout. In case (1), the outcome is a completely new DAG; in case (2), the new DAG must explicitly encode cross-turn dependencies through the task identifiers from the previous DAG; and in case (3), the resulting DAG reduces to a partial subgraph of the original DAG. We adopt the final data generation format introduced in ToolRL \cite{qian2025toolrlrewardtoollearning}. An example of a generated data sample is shown in Figure~\ref{sample}.

The quality of the data, particularly the synthetic data, is critical. To ensure reliability, we incorporate a rule-based verification mechanism into the generation pipeline. All tools, plan DAGs, tool calls, and observations produced by the LLM are required to be in JSON format. The first verification layer checks JSON validity. For plan DAGs, we apply AST \cite{patil2025bfcl} matching at each node to guarantee that the LLM only references the provided tools with the correct argument names. We further validate symbolic arguments by comparing each referenced key against the JSON schema of the predecessor’s output. Every tool call is verified against the plan DAG and the preceding tool call observations, ensuring adherence to the plan and the correct use of return values as inputs for dependent calls. Likewise, each observation is matched to its corresponding tool call, and its return value is checked against the tool’s JSON schema. If any verification step fails during generation, the LLM is required to restart the process. 

Our pipeline enables the generation of diverse synthetic queries derived from high-quality real APIs, each requiring resolution through a plan DAG. The difficulty of the queries is controlled by the topological structure of randomly generated DAG templates. Since these templates vary in structure, a fixed workflow for synthetic data generation is impractical. Instead, the graph-based pipeline provides a flexible end-to-end framework for producing such data. Finally, the rule-based verification mechanism ensures reliability: it not only checks JSON validity to guarantee compatibility with downstream benchmarks, but also leverages AST matching to validate the correctness of DAG instantiations during data generation.

\begin{tcolorbox}[colback=blue!5!white, colframe=blue!75!black, halign=flush left, title= Synthetic Data Sample]
\label{sample}
\textbf{System Prompt}: \newline 
You are a dialogue assistant designed to leverage tool calls to solve user tasks and provide structured responses.
\newline 
\newline 
{Available Tools}
\newline
In your response, you can use the following tools:
\{\{Tool List\}\}
\newline
\newline
{Steps for each turn}
\newline 
1. {Think}: Retrieve the relevant context and evaluate the current tool. \newline 
2. {DAG}: Produce a task list defined here \ref{task_list} \newline 
3. {Respond}: If a response is needed, generate one while maintaining consistency across user queries. \newline
\rule{\linewidth}{0.4pt} 
\newline
\textbf{Synethetic User Query}: ... \newline 
\rule{\linewidth}{0.4pt} 
\newline 
\textcolor{red}{</think>} \dots \textcolor{red}{</think>} [The think block is absent in the synthetic data but included during the training stage.]
\newline
\rule{\linewidth}{0.4pt} 
\newline 
\textcolor{green}{</DAG>} real DAG generated from pipeline \textcolor{green}{</DAG>}
\newline 
\rule{\linewidth}{0.4pt} 
\newline
<tool\_call> tool call 1st DAG layer </tool\_call> \newline
<obs> observation 1st DAG layer </obs> \newline 
... \newline 
<tool\_call> tool call last DAG layer </tool\_call> \newline 
<obs> observation last DAG layer </obs> \newline 
\textcolor{purple}{</response>} ... \textcolor{purple}{</response>} \newline 
\rule{\linewidth}{0.4pt} \newline 
\textbf{New User Query}: ... \newline 
\rule{\linewidth}{0.4pt} 
\textcolor{red}{</think>} \dots \textcolor{red}{</think>} \newline 
 \rule{\linewidth}{0.4pt} \newline
\textcolor{green}{</DAG>} new DAG based on the three scenarios defined in the multi-turn settings \ref{multiturn}. \textcolor{green}{</DAG>}\newline 
 \rule{\linewidth}{0.4pt} \newline
 ...
\end{tcolorbox}

\subsection{OrchDAG – Graph-based Reward Derived from the DAG for RLVR training}
\label{reward}

Due to the intricate tool interaction structure inherent in the synthetic data, the format reward, correctness reward, and parameter matching reward defined in \cite{qian2025toolrlrewardtoollearning} may remain sparse, even when initiating a large number of rollouts. Moreover, this reward does not account for structural dependencies among tools; thus, no reward is given when the LLM correctly predicts partial dependencies for these tools. 

To account for structural dependencies, and given that we have access to the ground-truth DAG during synthetic data generation, following \cite{Liu2025}, we use a weighted Graph Edit Distance (GED) as the reward signal at each turn. GED \cite{10.5555/2736754.2736796} measures the distance between two graphs by applying operations such as edge deletion, edge insertion, node insertion, or node relabeling to transform one graph into an isomorphic form of the other.

We define the reward for each turn as $$\text{R}_{\text{Total}} = \text{R}_{\text{Format}} + \alpha\text{R}_{\text{DAG}},\ \ \text{where}\ \ \text{R}_{\text{DAG}} = 1 - \frac{\text{GED}(\text{g}_1, \text{g}_2)}{\text{GED}(\text{g}_1, \emptyset) + \text{GED}(\text{g}_2, \emptyset)}$$
Here $\text{g}_1$ is the predicted DAG, $\text{g}_2$ is the ground-truth DAG. We define the following node equivalence when calculating GED: the tool name, parameter names, and parameter values are treated as a single unit, and equivalence is evaluated at the level of the entire tool call. The reward $\text{R}_{\text{Format}}$ is assigned a value of 1 if the output contains the special tokens in the correct order, and 0 otherwise. $\alpha$ is the hyperparameter used to balance the two types of rewards. In this reward design, we not only provide credit to LLMs for partially correcting the path during rollouts, but also make the rewards denser compared to the previous design. The multi-turn setting is naturally supported, as the ground-truth DAG is available at each turn.

\section{EXPERIMENTS}

In this section, we describe the generated dataset, focusing on the distribution of topological difficulty in the DAG templates and the proportion of single-turn versus multi-turn settings. For simplicity, we restrict evaluation to the two-turn setting, leaving extensions to longer horizons for future work. To ensure independence between training and test data, we construct them from two disjoint sets of data points from \textsc{APIGen} \cite{liu2024apigenautomatedpipelinegenerating} and \textsc{ToolACE} \cite{liu2025toolacewinningpointsllm}., as described in \ref{datapoint}. 

\begin{table}[ht]
\centering
  \caption{Synthetic Data Distribution (Height and width are hyperparameters controlling DAG complexity, and success rate is the proportion of data that passes rule-based validation)}
  \label{tab:freq0}
  \resizebox{0.8\linewidth}{!}{%
  \begin{tabular}{cccccc}
    \toprule
    Type & data \# & Multi-turn proportion & Average Height & Average Width & Sucessful Rate \\
    \midrule
    Training &1800& 30\% & 2.50 $\pm$ 0.12 & 3.4 $\pm$ 0.24 & 0.6\\
    Test &250& 25\% & 2.7 $\pm$ 0.08 & 3.1 $\pm$ 0.14 & 0.7\\
    \bottomrule
  \end{tabular}%
  }
\end{table}

\subsection{Task Difficulty}
We investigate two central questions: (1) Given the designed difficulty of our synthetic data, is the dataset solvable in principle? A dataset that cannot be solved even by advanced closed-source models such as Claude 4 or GPT-4o would lack practical utility. (2) If it is solvable, does the dataset offer a sufficient level of challenge to meaningfully evaluate model performance? To begin with, we evaluate several closed-source and open-source models by providing them with the system prompt defined in Section~\ref{sample}, and measure whether they can correctly predict the DAG by analyzing tool dependencies. In the multi-turn setting, models must generate the DAG by considering both the available tools and the observations from previous turns. We use Accuracy (pass@1) to evaluate performance, as our focus here is on assessing the task difficulty introduced by the dataset.

\begin{table}[ht]
\centering
  \caption{Pass@1 Accuracy for predicting the DAG for single/multi-turn settings (All experiments were run 10 times with the temperature of the base LLMs set to $0.1$)}
  \label{tab:freq1}
  \resizebox{0.6\linewidth}{!}{%
  \begin{tabular}{cccc}
    \toprule
    Models & Zero Shot & One shot & Three shots \\
    \midrule
    GPT-4o & \textbf{(0.18 $\pm$ 0.03)} & (0.22 $\pm$ 0.02) & \textbf{(0.24 $\pm$ 0.04)}\\
    Claude 4  & (0.15 $\pm$ 0.01) & \textbf{(0.23 $\pm$ 0.03)} & (0.22 $\pm$ 0.01)\\
    Claude 3.7 & (0.16 $\pm$ 0.04) & (0.18 $\pm$ 0.03) & (0.23 $\pm$ 0.01) \\
    Claude 3.5 & (0.08 $\pm$ 0.02) & (0.09 $\pm$ 0.03) & (0.08 $\pm$ 0.03)\\
    DeepSeek-R1 & (0.12 $\pm$ 0.02) & (0.14 $\pm$ 0.01) & (0.11 $\pm$ 0.04)\\
    Qwen2.5 3B & (0 $\pm$ 0) & (0 $\pm$ 0) & (0.02 $\pm$ 0.01) \\
    Qwen2.5 7B & (0.02 $\pm$ 0.01) & (0.03 $\pm$ 0.02) & (0.03 $\pm$ 0.02)\\
    \bottomrule
  \end{tabular}%
  }
\end{table}

For completeness, we also report Qwen3 pass@64 accuracy: 20.23\% for Qwen3-4B and 26.55\% for Qwen3-8B. From Table \ref{tab:freq1}, we can see that GPT-4o maintains the highest accuracy for the zero-shot setting and the three-shots setting with nearly same performance with Claude 4 in the one-shot setting. The accuracy for GPT-4o and also for Claude 4 shows that our dataset is solvable however the perofmrance for Claude 3.5, Qwen2.5 3B with accuracy 0, and Qwen 2.5 7B demonstrates the challenge of our datasets to current LLMs. Comparison between the one-shot and three-shot results shows that providing additional examples does not necessarily improve LLM performance in DAG prediction.

\subsection{Analysis of the Graph-based Reward Shaping in OrchDAG}
As discussed in Section~\ref{reward}, the reward signal from ToolRL \cite{qian2025toolrlrewardtoollearning} can be sparse in our data., which makes it difficult for reinforcement learning algorithms such as GRPO \cite{shao2024deepseekmathpushinglimitsmathematical}, DAPO \cite{yu2025dapoopensourcellmreinforcement}, and GiGPO \cite{feng2025groupingrouppolicyoptimizationllm} to efficiently improve the policy LLM, even with large rollouts. We apply ToolRL on the training single-turn dataset and evaluate it on the test set in the single-turn setting. In this setup, we train Qwen2.5 with GRPO using ToolRL, and convert the ToolCalls it generates into predicted DAGs, since the outputs follow a standardized JSON format. We use 8×100 A100 GPUs with Verl \cite{Sheng_2025} to complete the training. We evaluate performance using two metrics: Accuracy/step and Accuracy/user\_query. Accuracy/step measures correctness at the step level, where each individual action in a turn is assessed independently; a step may be correct even if the final tool execution graph is incorrect. Accuracy/user\_query measures correctness at the full query level, requiring the entire tool execution graph to be correct. 

\begin{table}[ht]
\centering
  \caption{ToolRL Performance on OrchDAG Single-turn Test Dataset
(The definitions of fine-grained and coarse rewards are given in ToolRL}
  \label{tab:freq2}
  \resizebox{0.4\linewidth}{!}{%
  \begin{tabular}{ccc}
    \toprule
    Model (Qwen2.5) & Acc/step & Acc/user\_query \\
    \midrule
    3B Coarse &0.517& 0 \\
    3B Finegrained &0.540& 0 \\
    7B Coarse & \textbf{0.609} & 0  \\
    7B Finegrained &0.594& 0  \\
    \bottomrule
  \end{tabular}%
  }
\end{table}
From Table~\ref{tab:freq2}, we observe that ToolRl performs reasonably well on certain steps within a single turn; however, it struggles to maintain a coherent overview of the entire execution. In contrast, as shown in Table~\ref{tab:freq1}, Qwen2.5-7B achieves 2\% accuracy in predicting the DAG. This indicates that for complex tool executions, it may be advantageous to first establish a high-level plan, such as a DAG, to guide the subsequent execution. We subsequently fine-tune Qwen2.5 using GRPO on the training single-turn dataset, guided by the proposed graph-based reward. We evaluate different hyperparameter settings: the use of entropy regularization and the KL loss, the choice of rollout number, and the number of training steps for the optimizer.

\begin{table}[ht]
\centering
  \caption{Performance of Graph-Based Reward on OrchDAG Single-turn Test Dataset (We report results using Acc/user\_query as the evaluation metric. The columns indicate the number of training steps, and $n$ denotes the rollout number.)}
  \label{tab:freq3}
  \resizebox{0.6\linewidth}{!}{%
  \begin{tabular}{cccccc}
    \toprule
    Model (Qwen2.5) / {Steps} & 15  & 30 & 45 & 60 \\
    \midrule
    3B KL n=4 & 0 & - & - & - \\
    7B n=4 & 0.184& 0.253& 0.241 & - \\
    7B KL n=4 & 0.184& 0.276& 0.276 & - \\
    7B KL Entropy n=4 & 0.195& 0.276& 0.253 & - \\
    7B KL n=8 & \textbf{0.23} & \textbf{0.33} & \textbf{0.402} & \textbf{0.391} \\
    
    \bottomrule
  \end{tabular}%
  }
\end{table}
In Table \ref{tab:freq3}, we observe that model size has a clear impact on performance. Moreover, the rollout number plays a crucial role, consistent with the intuition that larger rollout numbers enable greater exploration \cite{srivastava2025technicalsurveyreinforcementlearning}, thereby increasing the likelihood of reaching the correct DAG. To evaluate the effectiveness of the GED-based reward design, we conduct an ablation study using a coarser reward: the predicted DAG receives a reward of 1 if it exactly matches the ground-truth DAG, and 0 otherwise. Using the 7B model with KL and $n = 8$, the accuracy remains 0 even after 15 training steps. 

We then extend our experiments to the multi-turn setting. In this setting, we train Qwen2.5 on the entire training set with GRPO, using both the information from previous turns and the new user query as input, and evaluate performance on the full OrchDAG test dataset. Based on Table \ref{tab:freq3}, we report performance only at the 45th training step.

\begin{table}[ht]
\centering
  \caption{Performance of Graph-Based Reward on OrchDAG Single/Multi-turn Test Dataset (We report results using Acc/user\_query as the evaluation metric. The columns indicate the three multi-turn scenarios defined by \ref{fig:image2})}
  \label{tab:freq4}
  \resizebox{0.6\linewidth}{!}{%
  \begin{tabular}{cccccc}
    \toprule
    Model (Qwen2.5) / {Steps} & scenario 1  & scenario 2 & scenario 3\\
    \midrule
    7B KL Entropy n=4 & 0.112& 0.125& 0.218 \\
    7B KL n=8 & \textbf{0.156} & \textbf{0.203} & \textbf{0.352} \\
    
    \bottomrule
  \end{tabular}%
  }
\end{table}

From Table \ref{tab:freq4}, we observe that in the multi-turn setting, performance decreases across both experimental setups. The largest drops occur in Scenario 1 (tool-calling error) and Scenario 2 (requiring information from the previous turn), since these tasks depend not only on the new user query and the system prompt but also on information carried over from earlier turns. In contrast, the drop in Scenario 3 is smaller, as the new user query is completely independent of prior turns. To demonstrate generalizability, we further evaluate the trained model on StableToolBench \cite{guo2024stabletoolbench}, measuring solvable pass rates across L1, L2, and L3 categories. A task is considered successful when the predicted DAG matches the ground truth. We choose StableToolBench for its inherent complexity in tool interactions. In StableToolBench, GPT-4-0613 (CoT) achieves solvable pass rates of 45.5 (L1 instruction), 57.4 (L1 category), 48.8 (L1 tool), 43.0 (L2 instruction), 46.5 (L2 category), and 48.1 (L3 instruction). Under the same evaluation, our model attains 47.1, 56.4, 47.2, 41.3, 44.8, and 50.7, respectively.
\subsection{Training Insight Analysis}

As shown in Table \ref{tab:freq3}, performance generally improves as the number of training steps increases. However, when we extend training beyond this range, we observe a significant drop in performance around step 51. Inspired by DAPO \cite{yu2025dapoopensourcellmreinforcement}, We hypothesize that the performance drop may be caused by a low-entropy situation, where the model becomes overly confident and thus fails to explore sufficiently.
\begin{figure}[h]
  \centering
  \includegraphics[width=\linewidth]{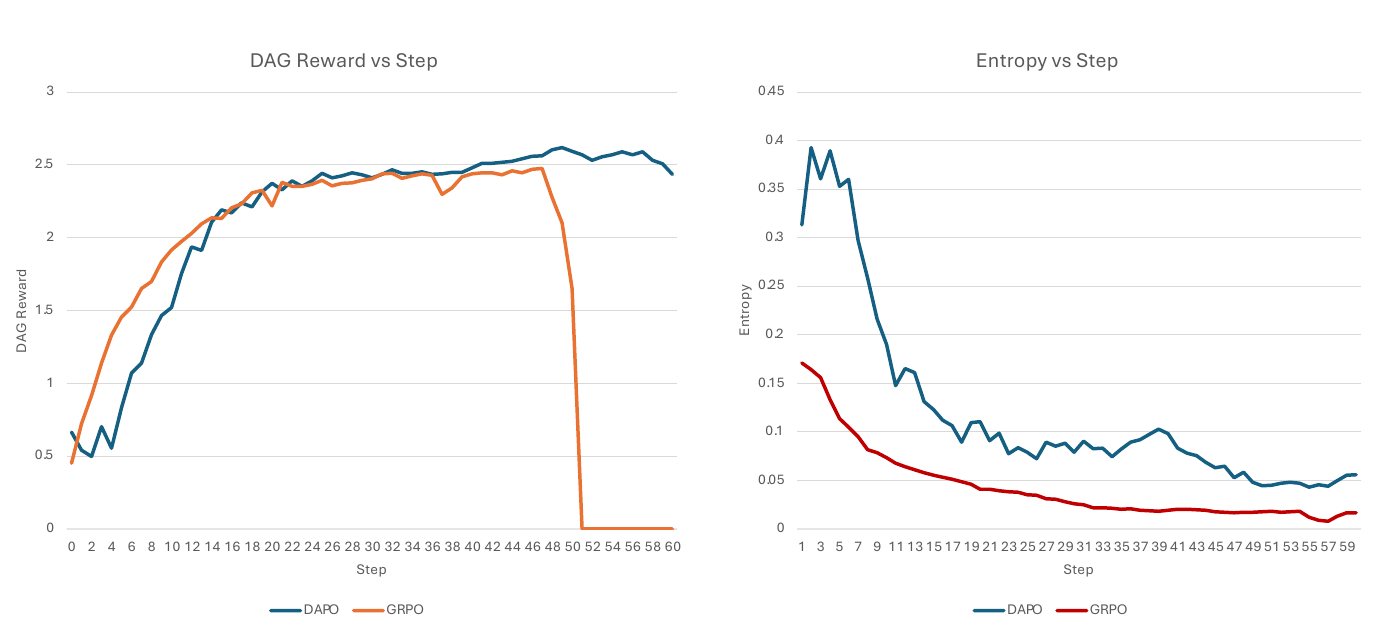} 
  \caption{Performance and Entropy Analysis with DAPO and GRPO}
  \label{fig:image5}
\end{figure}

After applying DAPO, we observe that the performance collapse no longer occurs, and the entropy remains at a relatively higher level in the later stages of training.

\section{CONCLUSION}
In summary, we introduce OrchDAG, a synthetic multi-turn data generation pipeline that models tool execution as DAGs with controllable complexity. Leveraging this dataset, we evaluate model performance and propose a graph-based reward to enhance RLVR training. Extensive experiments validate the effectiveness of our approach, underscoring the importance of exploiting the topological structure of tool execution graphs and managing data complexity in multi-turn tool use.

Nonetheless, our method remains limited in that it does not yet address multi-turn scenarios involving implicit dependencies, such as file operations in computer-use tasks. In future work, we aim to extend our framework to capture these implicit dependency cases.

\newpage
\bibliography{nips}
\bibliographystyle{plain}

\end{document}